\documentclass[11pt]{article}

\usepackage{authblk}
\usepackage[utf8]{inputenc} %
\usepackage[T1]{fontenc}    %
\usepackage{hyperref}       %
\usepackage{url}            %
\usepackage{booktabs}       %
\usepackage{amsfonts}       %
\usepackage{nicefrac}       %
\usepackage{microtype}      %
\usepackage[table]{xcolor}         %
\usepackage{amsmath}
\usepackage{cleveref}
\usepackage{diagbox}
\usepackage{fullpage}

\usepackage{graphicx}           %
\usepackage{comment}
\usepackage{tabularx}

\usepackage{subcaption}

\usepackage{algorithm}
\usepackage[noend]{algpseudocode}

\algrenewcommand\algorithmicrequire{\textbf{Initialize}}
\algrenewcommand\algorithmicensure{\textbf{Input}}
\usepackage{enumitem}

\newcolumntype{L}{>{\raggedright\arraybackslash}X}

\newcommand{\ourflip}{PoolFlip}

\author[1]{Xavier~Cadet}
\author[2]{Simona~Boboila}
\author[1]{Sie~Hendrata~Dharmawan} 
\author[2]{Alina~Oprea}
\author[1]{Peter~Chin}

\affil[1]{Dartmouth College, Hanover, NH 03755, USA}
\affil[2]{Northeastern University, Boston, MA 02115, USA}

\title{\ourflip{}: A Multi-Agent Reinforcement Learning Security Environment for Cyber Defense}

\date{} %

\begin{document}

\maketitle

\begin{abstract}
Cyber defense requires automating defensive decision-making
under stealthy, deceptive, and continuously evolving adversarial strategies.
The FlipIt game provides a foundational framework for modeling interactions between a defender and an advanced adversary that compromises a system without being immediately detected. In FlipIt, the attacker and defender compete to control a shared resource by performing a Flip action and paying a cost. However, the existing FlipIt frameworks rely on a small number of heuristics or specialized learning techniques, which can lead to brittleness and the inability to adapt to new attacks. 
To address these limitations, we introduce \ourflip{}, 
a multi-agent gym environment that extends
the FlipIt game to allow efficient learning for attackers and defenders. Furthermore, we propose Flip-PSRO, a multi-agent reinforcement learning (MARL) approach that leverages population-based training to train defender agents equipped to generalize against a range of unknown, potentially adaptive opponents. Our empirical results suggest that Flip-PSRO defenders are $2\times$ more effective than baselines to generalize to a heuristic attack not exposed in training. In addition, our newly designed ownership-based utility functions ensure that Flip-PSRO defenders maintain a high level of control while optimizing performance.
\end{abstract}

\section{Introduction} \label{section:introduction}

The number and complexity of cyber threats are expected to grow alongside advances in Artificial Intelligence (AI).
While these developments present new challenges, AI also offers opportunities to enhance cybersecurity through techniques such as anomaly detection~\cite{GAMAGE2020,LeungAnomaly2005,Beehive}, process automation~\cite{subudhi2024effectiveness}, and adaptive defense mechanism~\cite{salem2024advancing}. 
One promising approach to cybersecurity automation is the use of Reinforcement Learning (RL) to develop Automated Defense Systems (ADS) capable of responding to evolving threats such as Advanced Persistent Threats (APTs)~\cite{apruzzese2022cyber,marl_cybersecurity2025,hammar2024optimal,mcdonald2024,nguyen2021deep,wiebe2023}. 
RL has typically been studied in simulated environments, which facilitate experimentation and large-scale data generation.
Cybersecurity games such as CybORG~\cite{cyborg_acd_2021}, CyberBattleSim~\cite{CyberBattleSim} and FlipIt~\cite{vandijkFlipItGameStealthy2013} provide valuable abstraction among simulation-based approaches.
While CybORG and CyberBattleSim require significant domain expertise to simulate realistic network topologies including servers, processes, and network events, FlipIt, the focus of this study, abstracts away the underlying network and systems components to facilitate game-theoretic modeling. 

The FlipIt game, introduced by van Dijk et al. \cite{vandijkFlipItGameStealthy2013}, models stealthy takeovers in cybersecurity as a two-player partially observable game, where acquisition of a resource incurs a cost and resource possession presents a reward. Compared to other security games, in FlipIt the attacker and defender compromise the resource without being immediately detected, which models interaction with stealthy adversaries, such as advanced persistent threats (APTs). Laszka et al. \cite{laszkaFlipThem2014} extended this framework to multi-resource settings, and Pham and Cid \cite{phamAreWeCompromised2012} considered adding a resource checking action. These extensions allow for further analysis of more realistic cybersecurity configurations, but they require hand-crafted designs tailored to specific opponent behaviors.
Subsequent studies—such as Oakley et al. \cite{oakleyQFlipAdaptiveReinforcement2019} and Greige et al.~\cite{greigeDeepReinforcementLearning2022a}—have explored reinforcement learning (RL) techniques, including Q-Learning \cite{watkinsQlearning1992} and Deep Q-Networks (DQN) \cite{mnihPlayingAtariDeep2013}, to learn player strategies.
However, these works focus on the online learning paradigm, where an agent learns through prolonged exposure to a fixed opponent.  Although this demonstrates the agent’s ability to adapt over time, it does not reflect more realistic deployment scenarios where a defender must generalize their strategies against new opponents, potentially sampled from a known distribution.
For example, an intrusion detection system must immediately identify and block new attack variants from known malware families, without a chance to ``learn'' the specific attack pattern during the actual attack.
In contrast to prior FlipIt studies, we explore the training of defender blue agents equipped to generalize and succeed against a range of unknown, potentially adaptive opponents at deployment time, better reflecting the real-world demands of proactive cyber defense.

In this work, we design a multi-agent gym environment, \ourflip{}~\footnote{\ourflip{} is publicly available at \url{https://github.com/xcadet/poolflip}}, that extends the FlipIt cybersecurity game to allow efficient learning for both attackers and defenders.  
Compared to prior work~\cite{vandijkFlipItGameStealthy2013,greigeDeepReinforcementLearning2022a,oakleyQFlipAdaptiveReinforcement2019}, we extend this environment with new parameterized heuristic agents and trainable agents that can act as both defenders and attackers and are representative of realistic attack patterns. The action space is extended with the Check action to detect the state of a given resource and guide the strategy. 

Furthermore, we introduce a MARL approach, named Flip-PSRO, to learn optimized policies through population-based training, similar to recent successful implementations in strategy games like StarCraft~\cite{vinyalsStarCraft2019}, Go~\cite{Silver_2016}, and Barrage Stratego~\cite{mcaleer2021pipelinepsroscalableapproach}. Flip-PSRO adopts population-based training in MARL~\cite{lanctotUnifiedGameTheoreticApproach2017} and Proximal Policy Optimization (PPO)~\cite{schulman2017proximal} to iteratively learn the best response to the opponent’s mixed policies.
The PSRO algorithm finds an approximate Nash equilibrium over the population by solving a meta-game, which serves as the selection mechanism for the mixed strategy (probability distribution over policies).
Unlike strategy games that utilize a typical win-loss utility metric for solving the meta-game, we design new utility targets tailored to the setting of stealthy takeover games, such as the win rate by ownership and specialization ratio.

We conducted extensive experiments in \ourflip{} to evaluate multiple defensive strategies against a diverse set of adversarial behaviors. The experiments show that Flip-PSRO achieves higher robustness and generalization against varied opponents compared to several baseline methods. For example, Flip-PSRO's average reward of 31 over the heuristic pool exceeds our best performing heuristic (13.8) and the Iterated Best Response method (23) showing better generalization over the training pool. Flip-PSRO also exhibits good transfer quality to unseen opponents (opponent strategies that were not present in training), achieving a $2\times$ higher reward (of 32) on average compared to the Iterated Best Response method (reward of 14). In addition, our experiments demonstrate the benefit of the new ownership-based utility target for the meta-game, which ensures that Flip-PSRO can maintain a high level of control while optimizing performance. 

\section{Related Work and Background} 
\label{section:background}

\noindent \textbf{FlipIt and Extensions:} FlipIt~\cite{vandijkFlipItGameStealthy2013} is a security game aimed at encouraging research in cyber defense. It was originally formulated as a non-zero-sum cybersecurity framework where two players compete for control over a shared resource. In their seminal paper, van Dijk et al.~\cite{vandijkFlipItGameStealthy2013} proposed several non-adaptive heuristic strategies and proved results about their strongly dominant opponent's Nash Equilibrium, and a Greedy adaptive strategy, proven to not always be optimal.
Theoretic extensions of FlipIt include multiple resources ("FlipThem"~\cite{laszkaFlipThem2014}, and "FlipNet"~\cite{saha2017flipnet}), information leakage ("FlipLeakage"~\cite{Farhang2016FlipLeakage}), and insiders~\cite{Hu2015FLipitInsiders}.

The exploration of deep-RL techniques in FlipIt has been limited. We note the works of Oakley et al.~\cite{oakleyQFlipAdaptiveReinforcement2019} on Q-Learning paradigms in this setting, and Greige et al.~\cite{greigeDeepReinforcementLearning2022a} on the applications of Deep Q-Networks to automated defenses. However, none of these papers study the transferability of their defense strategies to more diverse adversarial settings. We improve upon these prior works by exploring more adaptive learning methods under the Policy-Space Response Oracle paradigm, and show empirically that our defense methods can successfully respond to an extended set of evolving attackers.

\vspace{3pt} \noindent \textbf{Deep RL and its applications:} Deep RL enables AI agents to learn complex tasks by interacting with an environment in single-agent and multi-agent settings~\cite{smithLearningPlayAny2021}, finding successful applications in complex sequential decision-making games such as Atari~\cite{mnihHumanlevelControlDeep2015}, Go~\cite{Silver_2016} and StarCraft~\cite{vinyalsStarCraft2019}.
Recently, there has been an increasing interest in automating cyber defense strategies using RL-based agents~\cite{apruzzese2022cyber,marl_cybersecurity2025,hammar2024optimal,mcdonald2024,nguyen2021deep,Singh:2024:HMARL,wiebe2023}, leading to the development of RL cybersecurity environments such as CybORG~\cite{cyborg_acd_2021} and CyberBattleSim~\cite{CyberBattleSim}. These environments simulate realistic networks consisting of multiple hosts, processes, and network connections, in contrast to FlipIt, where the underlying network is abstracted away to facilitate game-theoretic modeling. 

\vspace{3pt} \noindent \textbf{Policy-Space Response Oracles (PSRO):}
A recent development in multi-agent RL is the Policy-Space Response Oracles framework proposed by Lanctot et al. \cite{lanctotUnifiedGameTheoreticApproach2017}, which uses deep RL algorithms to calculate a best response policy against multiple opponent strategies.
Bighashdel et al.~\cite{Bighashdel2024} provide a comprehensive survey on the PSRO algorithm and its application in various domains such as sequential auctions~\cite{Zhang_psro_auctions}, green security~\cite{Wang2019greensecurity,xu2021robustreinforcementlearningminimax}, robust RL~\cite{liang2024gametheoreticrobustreinforcementlearning}, cybersecurity~\cite{guo2023patrol} and large-scale games~\cite{zhou2022efficientpolicyspaceresponse}. Notably, PSRO has reached state-of-the-art performance in strategy games like StarCraft~\cite{vinyalsStarCraft2019} and Barrage Stratego~\cite{mcaleer2021pipelinepsroscalableapproach}, where it outperformed human experts and other AI methods. 

Next we present the building blocks of the PSRO framework and its general formulation. The \textbf{normal form} is a common representation, where a game is denoted as $G=(N, \Pi, U)$~\cite{zhang2025surveyselfplaymethodsreinforcement}, where $N=\{1, 2, \cdots, n\}$ are the players, $\Pi=\{\Pi_1, \Pi_2, \cdots, \Pi_n\}$ is the pure strategy space of all players and $U=\{u_1, u_2, \cdots, u_n\}$ are the utility functions of the players, with $u_i: \Pi \rightarrow \mathbb{R}$ being the \textbf{utility} function for player $i$. The utility matrix, also called the payoff matrix or the evaluation matrix, captures the outcomes of the game when different strategies are played against each other.

A \textbf{pure strategy} defines a specific action for a player in a game, while a \textbf{mixed strategy} of the player $i$ defines a probability distribution over the set of pure strategies of $i$, and is represented as $\sigma_i \in \Delta(\Pi_i)$, where $\Delta$ is a probability simplex.  Furthermore, a \textbf{strategy profile} consists of a vector of strategies $\pi=(\pi_1, \pi_2, \cdots, \pi_n)$ for all players.

Let $\pi_i$ be the strategy of the player $i$, and $\pi_{-i}$ the joint strategy profile of all players other than $i$. The \textbf{best response} of the player $i$ is the strategy that maximizes the payoff of the player $i$':
\begin{equation}
BR_i = \text{arg } \text{max}_{\pi_i} u_i(\pi_i, \pi_{-i})
\end{equation}

A \textbf{restricted game} is a projection of the full game $G$. The players choose from a restricted strategy set $S \subseteq \Pi$, and the utilities can be approximated by simulation. Deep RL techniques (e.g., PPO, DQN) are used to compute an \textbf{approximate best response}. 

PSRO aims at handling large-scale games in which game-theoretic analysis is hindered by the wide number of strategies available.
To this end, PSRO considers a \textbf{meta-game}, a restricted game that uses a tractable subset of strategies while still approximating the full game well. The utilities for the strategy profiles are estimated by simulation.
PSRO introduces the concept of a \textbf{meta-strategy solver (MSS)}, which selects a profile $\sigma \in \Delta(S)$ from the restricted strategy set $S$, where $\Delta$ is a probability simplex. A common MSS is the uniform distribution, where the next opponent strategy is chosen uniformly from the restricted set. Other meta-strategy solvers can be defined, based on the next best-response target, or the response objective. 

\vspace{3pt} \noindent \textbf{PSRO in cybersecurity:}
The application of PSRO and other population-based training methods to traditional cybersecurity domains like intrusion detection, malware analysis or network defense is still in the beginning. We note the work of Ma et al.~\cite{ma2024evolvingdiverseredteamlanguage}, who apply meta-game analysis to AI safety and red teaming. Guo et al.~\cite{guo2023patrol} train joint policies that are robust against adversarial attacks. Furthermore, Tong et al.~\cite{Tong_Laszka_Yan_Zhang_Vorobeychik_2020} use a double-oracle framework to compute a policy that prioritizes alerts in intrusion detection systems, while
Cui et al.~\cite{ICLR23-Cui} conduct population-based training to detect cache timing attacks. 

In this paper, we apply population-based training to a core cybersecurity problem: protecting critical assets from malicious actors. \ourflip{} is the first framework to explore PSRO in the context of the FlipIt security game, modeling a diverse population of attack and defense strategies in an evolving threat landscape.

\section{The \ourflip{} Framework}
\label{section:game}

We introduce \textbf{\ourflip}, an extension of the FlipIt cybersecurity game \cite{vandijkFlipItGameStealthy2013}, designed as a Multi-Agent Reinforcement Learning (MARL) framework. The game has three components: the system resource(s), an attacker, and a defender. The target of the threat is the system resource, which can be any critical asset, such as a password, software, server, network, etc.
The objective of both the attacker and the defender is to control the resource: the attacker attempts to compromise it, while the defender's task is to minimize damage by recapturing it.
We distinguish between players and strategies. Although a defender faces a single attacker during a game, there can be multiple types of attack strategies and defense strategies.

Under the RL paradigm, \ourflip{} is formulated as a partially observable Markov decision process (POMDP)~\cite{Oliehoek2016ACI} comprising observation space $O$, state space $S$, action space $A$, transition probability $P$, and reward function $r$. An agent with incomplete observations interacts with the environment, taking actions that lead to state transitions and receiving rewards, with the goal of maximizing its cumulative reward (or discounted return) over time. 

\subsection{Actions}  
In the original FlipIt game, there are two actions: to \textbf{Flip} a resource or \textbf{Not to Flip} the resource, the latter of which we refer to as \textbf{Sleep}. \ourflip{} uses an extra action called \textbf{Check}, which allows an agent to inspect the state of a resource without changing its ownership.
Although both attackers (Red Agent) and defenders (Blue Agent) use the same actions, the implications differ for each type of agent.

\vspace{2pt}\noindent\textbf{Sleep:}  
Global action, which does not target any resource. The agent is inactive.  

\vspace{2pt}\noindent\textbf{Check:}  
Targeted action. The agent probes a specific resource to learn about its current ownership status and the timing of any recent flips.
\begin{itemize}
    \item \emph{Blue Agent:} Examples of check actions include performing targeted host scans, reviewing system logs, or querying security information and event management (SIEM) systems.
    \item \emph{Red Agent:} Attackers can perform stealthy reconnaissance such as fingerprinting a host, checking for open ports or services, or validating persistence, without alerting the defender.
\end{itemize}

\vspace{2pt}\noindent\textbf{Flip:}  
Targeted action. The agent attempts to take over a specific resource.
\begin{itemize}
    \item \emph{Blue Agent:} A Flip action represents active defense, such as removing malware, patching a system, resetting credentials, or restoring a host to a secure state.
    \item \emph{Red Agent:} Using Flip, the attacker attempts to take control of an asset by exploiting a vulnerability, deploying a payload or escalating privileges.
\end{itemize}

\vspace{2pt}\noindent\textbf{Action Space:}  
We structure the action space to reflect the presence of global and targeted actions, which leads to a total of $1 + 2 \times R$ actions:
[Sleep, Flip, Check, $\ldots$, Flip Resource $R$, Check Resource $R$], where $R$ is the total number of resources.

\subsection{Observations}

\ourflip{} models adversarial engagements where the Red and Blue Agents compete over control of multiple cyber assets, each with limited capacity to act and partial information about the opponent's strategy.
At each time step $t$, the agent receives an observation $o_t$ from the environment, which is used to update the agent's knowledge about the system. However, the agent's knowledge is imperfect and may become stale. Since the game is \emph{stealthy}, the players do not automatically find out when the other player has last moved. The agent can update its information about a resource only by calling Flip or Check on that resource.
Agents keep track of the following pieces of information:
\begin{itemize}
    \item Is the agent the current owner of the resource?
    \item The time since the agent captured the resource.
    \item The time since an opponent has captured the resource.
\end{itemize}

For trained agents, this information is pre-processed into a vector to be used with deep reinforcement learning algorithms. 
Given a memory limit $M$, and assuming that a player flipped the resource $\Delta t$ time steps ago, the encoded time is $min(\Delta t, M)$. We can control the number of time steps that the agent is tracking by varying $M$, which directly impacts the input size of the neural network. To feed this information to the neural network, we generate a one-hot encoded vector, namely $\{0, 1\}^{M}$; we also add an extra cell in the memory buffer to represent the unknown status. The size of the observation vector per resource is therefore $2+2M$.

For multiple resources, the observation vectors are concatenated, resulting in a vector of size $(2+2M) \times R$, where $R$ is the number of resources. Alternatively, we can view this vector as matrix $\mathbb{R}^{R \times (2+2M)}$ where the row $i$ contains information on the status of the $i$-th resource.

\subsection{Gain, Cost and Reward}
\ourflip{} is a strategy game in which players balance utility-related decisions with individual control of resources.

\vspace{3pt}\noindent\textbf{Resource Ownership:}
The resource ownership remains unchanged if: (1) no other player contests the resource (using Flip), or (2) the current owner contests it.
When multiple players claim the resource at the same time, a new owner is randomly selected from the contestants unless the previous owner is also a contestant. Hence, \ourflip{} transitions are deterministic except in the cases of simultaneous flips without the prior owner, which introduces stochasticity.
Let $C$ be the set of contestants that claim ownership of resource $i$, at step $t$, and $\texttt{Uniform}(C)$ be a uniform distribution over this set. The owner of $i$ at time step $t$, $O_i(t)$, is: 
\begin{align}
O_i(t) = 
    \begin{cases}
      O_i(t-1) & \text{if } C = \emptyset \text{ or }O_i (t-1) \in C\\
      X \sim \texttt{Uniform}(C) & \text{otherwise}
    \end{cases}
\label{eq:owner}
\end{align}

\vspace{3pt}\noindent\textbf{Gain:}
The gain quantifies the benefit of resource ownership for each player. 
In \ourflip, resources can have different gains, depending on their criticality; for example, compromising a database server is worth more to an attacker than compromising a user machine. Furthermore, the gain can be different for each player, depending on the utility of the resource towards achieving their end goal. 

\vspace{3pt}\noindent\textbf{Cost:}
Each action has a positive cost associated with it. In general, Flip is more expensive than Check; however, our framework can model different cost values for each resource and player. For the defender, flipping can represent resetting a machine to a clean state, which is more disruptive than monitoring and investigating alerts (i.e., Check). Similarly, for the attacker, exploiting a vulnerability to obtain root privileges (Flip) is more difficult than scanning the network (Check), hence the higher cost.

\vspace{3pt}\noindent\textbf{Reward:}
We compute the reward as the difference between the gain and the cost. Given a set of resources $S$ owned by agent $A$ at time step $t$, the reward is:
\begin{align}
\text{Reward} (A, t) = \sum_{i \in S} \text{Gain}(A, i) - \text{Cost}(A, a_t)
\label{eq:reward}
\end{align}
\noindent where $a_t$ is the action taken by the agent at the time step $t$.

We are interested in developing agents who can control the resource longer than their opponents while also achieving a positive reward.
A straightforward strategy to maintain control of a resource and maximize possession time is to flip on every turn. However, such a strategy can lead to negative rewards when cost offsets utility.
Consider $c = \text{Cost}(A, \text{Flip} (i))$, the cost to flip the resource $i$ by the agent $A$ and $g = \text{Gain}(A, i)$, the unit gain for $i$ within each ownership time step. If $c > g$, the agent must wait before flipping again for at least $c/g$ time steps to achieve a positive reward. Therefore, even against a passive opponent, an agent may receive negative rewards if it attempts to flip the resource too frequently.

\section{Heuristic Agents}
\label{sec:heuristic}

We consider two types of heuristic strategies (or agents): \textit{adaptive} and \textit{non-adaptive}.
Adaptive agents change their strategy based on the observations they obtain while playing. The non-adaptive agents follow a rule-based strategy that does not take into account changes in the environment. Table~\ref{tab:heuristic_agents} describes the various types of heuristic agents in detail, including examples of their application to cybersecurity.

\begin{table}[!htbp]
\caption{Heuristic Agents and some of their applications to cybersecurity.}
\small
    
\centering
\begin{tabularx}{\textwidth}{L|L|L|L}
\hline
\textbf{Agent}&\textbf{Behavior} & \textbf{Defense} & \textbf{Attack}\\
\hline
\multicolumn{4}{c}{\bf Non-adaptive Heuristic Agents} \\ \hline
\textit{SleepOnly} &Inactive. &Inactive or mis-configured defense. & Dormant attacker.\\ \hline
\textit{Periodic} &Flips at periodic intervals with an optional delay.& Heartbeats and key rotations, mechanisms that enforce security policies at predictable intervals.
& Cron jobs that execute malicious scripts, potentially synchronizing with external events. \\ \hline
\textit{Burst} &Periodic, multiple flips in bursts.& Heavy inspections in tight cycles before cooling down to conserve resources.
& Stealthy attacker that acts aggressively in short windows, then goes quiet to avoid detection. \\ \hline
\textit{Awakening} &Probability of flipping increases with time since last flip.& Defender whose suspicion grows with time. 
& Stealthy adversary whose likelihood to act increases with time. 
\\\hline
\multicolumn{4}{c}{\bf Adaptive Heuristic Agents} \\ \hline
\textit{Retaliating} &Adjusts flipping frequency based on the opponent's aggressiveness. & Defensive activity increases in response to detected threats and relaxes during quiet periods.
& Attacker intensifies its attempts to re-establish foothold after being evicted.\\\hline
\textit{Periodic Check} &Checks periodically, and flips during its next phase after detecting a  takeover. &Scheduled checks, periodic assessments followed by patch updates. &Cron jobs that execute malicious scripts in some system states. \\\hline 
\textit{Periodic Aggressive Check} &Checks periodically, and flips during its next turn after takeover detected. &Same as Periodic Check. & Same as Periodic Check. \\\hline
\end{tabularx}
\label{tab:heuristic_agents}
\vspace{-15pt}
\end{table}

\vspace{3pt}\noindent\textbf{Non-adaptive heuristic agents (observation-agnostic):} Non-adaptive heuristic strategies follow a fixed rule-based pattern that does not take state changes into account.

\begin{itemize}

\item \textit{SleepOnly:}
A passive baseline agent that never flips.

\item \textit{Periodic:} An agent that flips at regular fixed intervals with an optional delay. This agent is parameterized by a \emph{phase} (number of time steps separating two flips) and a \emph{delay} (number of time steps before the first flip). For example, with a delay of $2$ and a phase of $3$, we would observe the following pattern of Sleep ($S$) and Flip ($F$) actions during the first $10$ time steps: $[S, S, F, S, S, F, S, S, F, S]$, where the first two $S$ are due to the delay parameter.

\item \textit{Burst:}
An agent that acts in bursts after a period of inactivity.
This agent is parameterized by a \emph{phase}, a \emph{delay} (similar to the Periodic Agent), and a \emph{burst} parameter that dictates the number of consecutive flips. For example, with a delay of $2$, a phase of $3$ and a burst of $3$ we observe the following pattern over the first $10$ time steps: $[S, S, F, F, F, S, S, F, F, F]$.

\item \textit{Awakening:} 
An agent whose probability of flipping increases with the time since its last Flip based on its parameter $\lambda$. Specifically, the longer the agent sleeps (i.e., does not flip), the more likely it is to flip in the next step, following an exponential cumulative distribution function (CDF).

The probability of flipping at time $t$ since the last awakening is computed as $P(\text{flip at time } t) = 1 - e^{-\lambda t}$, starting at 0 when $t = 0$ and asymptotically approaching 1 as $ t \to \infty$, meaning the agent becomes increasingly likely to flip as time progresses.

\end{itemize}

\vspace{3pt}\noindent\textbf{Adaptive heuristic agents (observation-aware): }
Adaptive heuristic agents take advantage of the knowledge they acquire from observations to evolve their strategy.

\begin{itemize}
\item \textit{Retaliating:}
An agent that adapts how frequently it flips based on whether it is currently in control of the resource.
When the opponent is more aggressive, the agent increases its flipping frequency, and when the opponent is more passive, the agent decreases its frequency. Its shortest phases is $\max{\frac{1}{2} \times \text{phase}}$ and its maximal phase is its starting phase.

\item \textit{Periodic Check (PC):} This opponent uses the phase and delay parameters to perform checks to detect if a resource has been captured. If so, it flips in the following phase.

\item \textit{Periodic Aggressive Check (PAC):} This agent is more aggressive than the Periodic Check agent, flipping in the next turn after detecting a takeover (instead of waiting for an entire phase). One special case is PAC with a phase of $1$, namely checking on every turn and flipping as soon as it does not control the resources. In settings with low Check cost, this strategy is challenging to defeat, we name this variant UPAC.
  
\end{itemize}

As Table~\ref{tab:heuristic_agents} illustrates, both adaptive and non-adaptive heuristic agents find important applications in cybersecurity. On the defense side, periodic assessments or continuous automated monitoring are followed by mitigation actions such as patch updates, removal or suspicious processes and files, or system reset to a previous clean state. These security measures can be simulated with the different flavors of heuristic agents, including Periodic Check and PAC. On the attack side, stealthy adversaries often stay dormant for long periods of time to remain hidden, and may abuse the cron utility to perform task scheduling for initial or recurring execution of malicious code~\cite{cronjobs}. This type of adversarial behavior can be modeled with the parameterized heuristic agents studied here. 

\section{Learning Algorithm}

The design of effective learning algorithms for the training of defenders in the \ourflip{} environment presents several challenges, mainly due to partial observability and the need to generalize among various opponents.
We start by applying Proximal Policy Optimization (PPO) \cite{schulman2017proximal}, a widely used reinforcement learning  algorithm, to train the defenders against fixed heuristic red agents.
While PPO can successfully learn specialized policies —what we call \textit{Specialists}— these strategies typically fail to generalize beyond the specific red agent they were trained against, echoing findings in prior work \cite{kielyAutonomousAgentsCyber2023}.
One attempt at generalization is to learn a best policy against each heuristic red agent sequentially, using the Iterated Best Response (IBR) method. However, IBR can cause dramatic strategy shifts, as agents make large policy jumps that consider only locally optimal choice, without consideration for responses of future opponents~\cite{Roughgarden2010}.

To overcome this limitation, we move toward more general blue strategies trained against a distribution of red opponents.
To this end, we adapt the Policy-Space Response Oracles (PSRO) \cite{lanctotUnifiedGameTheoreticApproach2017} framework, which iteratively builds a pool of policies and computes best responses against mixtures of strategies.
Our variant, Flip-PSRO, introduces novel design choices for building the pool and selecting a mixture of opponents based on meta-strategy solvers adapted to our problem space. 

\subsection{Key Design Choices}

\noindent\textbf{Meta game payoff:} Unlike zero-sum games where one player's gain equals another player's loss, general-sum games like \ourflip{} have a more complex payoff structure. In \ourflip{} the reward depends both on the cost of an action and on the gain accumulated from the ownership of the resources. While strategy games like Poker or Starcraft use a typical win-loss utility metric for the meta-game, we design new utility targets, or response objectives, tailored to the setting of stealthy takeover games. The duration of resource control is one such example, leading to a new formulation of what win-loss represents in \ourflip{} (see Section~\ref{section:response_objectives}).

\vspace{3pt}\noindent\textbf{Pool diversity:}
The pool consists of rule-based strategies that capture domain knowledge about realistic defense and attack strategies, as described in Table~\ref{tab:heuristic_agents}. These heuristics are parameterized by phase, burst length and awakening coefficient (exponential), modeling a diverse range of behaviors. Thus, the pool contains different strategic archetypes: aggressive (Periodic, Burst), unpredictable (Awakening), balanced (Periodic Check).

\vspace{3pt}\noindent\textbf{Game symmetry:}
We consider a symmetric game where the two players, the defender and the attacker, utilize the same reward formulation and can execute an identical strategy. Therefore, our Flip-PSRO algorithm can evolve a single strategy that is applicable to either player. We leave as future work the modeling of games with asymmetric cost/gain, and meta-games over mixed strategies using different probability distributions for each player.

\subsection{Response Objectives} 
\label{section:response_objectives}
In this work, we introduce the following response objectives, constructed as utility functions that need to be maximized over the agent's strategy profile to derive a best response strategy. We operationalize the paper’s stated objectives (‘maintain control and optimize performance’) using the following response objectives: (i) win rate by ownership — ensuring long-term control; and (ii) normalized performance gap — ensuring competitive performance relative to specialists.

\vspace{3pt} 
\noindent \textbf{Win Rate by Ownership:} 
We define the \textit{ownership score} as the fraction of time during which an agent controls a specific resource. A defender who maintains a high ownership score is able to identify vulnerable assets and implement effective security measures to keep the system operational, while an attacker who has a high ownership has inflicted a successful disruption on the system. 
For a single resource $r$, the ownership score of the agent $i$ is:
\begin{equation}
\text{own}(i, r) = \frac{\sum_{t}^{T}\mathbf{1}_{r(t) = i}}{T}
\end{equation}
where $\mathbf{1}_{r(t) = i}$ indicates whether agent $i$ controls resource $r$ at time step $t$ and $T$ is the total number of time steps. An agent that controls the resource at every time step receives an ownership score of 1, while an agent that never controls it receives a score of 0. Intermediate values indicate partial or intermittent control.

A defender needs to control the resource for enough time to ensure availability to the users of the system. For example, a 50\% ownership ensures that the defender owns the resource half the time; however, it might not be sufficient in cases where a higher level of control is necessary. We denote by $t\%$ the required duration of ownership for the defender to be declared the winner.
The \emph{win rate by ownership} for the player $i$ represents the fraction of games won by $i$, using a predefined ownership target of $t\%$.

\vspace{3pt} 
\noindent \textbf{Normalized Performance Gap:}
Specialists are agents trained to maximize their reward against a particular type of heuristic opponent. We use the following notation: $H$ is a heuristic agent, and $S_H$ is a specialist trained against $H$. Since specialist agents $S_H$ have learned counter-measures against a specific attack pattern, they are a good approximation of the best response against agent $H$.

Let $r(S, H)$ be the average reward of the specialist agent against $H$, after training has converged. For any agent $i$ playing against opponent $H$, we define the \textit{performance gap} as:
$$
\text{Gap}(i, H) = \max(0, r(S, H) - r(i, H))
$$
This gap measures how far agent $i$ is from matching the performance of the specialist against $H$. Larger values indicate worse relative performance.

To make comparisons across opponents fair, we normalize the gaps across all heuristic types. Let $\text{Gap}_1, \dots, \text{Gap}_n$ be the gaps for each heuristic in a set of $n$ heuristics. The normalized gap for each is given by:
$$
\text{NormGap}(i, H_j) = \frac{\text{Gap}(i, H_j) - \min_k \text{Gap}(i, H_k)}{\max_k \text{Gap}(i, H_k) - \min_k \text{Gap}(i, H_k)}
$$

\subsection{Flip-PSRO algorithm} 

Algorithm~\ref{alg:psro} describes our iterative learning method, Flip-PSRO. The goal of this algorithm is to train a robust \emph{defender} that performs best against a pool of attackers. 
Deep RL, specifically PPO~\cite{schulman2017proximal}, represents the players' response oracle and is used to learn an optimal defender strategy.

\begin{algorithm}[t]
    \centering
    \caption{Flip-PSRO Algorithm}\label{alg:psro}
    \begin{algorithmic}[1]
     \Ensure Initial policy set $\Pi = \{$Heuristic Policies\};\newline
     $RO$ (Response Objective) $\in$ \{reward, win rate, normalized performance gap\};\newline
     $U^\Pi_{RO}$ is the expected utility matrix, initially empty. \newline
     Initialize $\sigma=\texttt{Uniform}(\Pi)$ \Comment Uniform is the MSS for the first iteration
     \For{iteration $t$}
       \For{episode $e$}
         \State Sample opponent policy $\pi' \gets \sigma $
         \State Train PPO policy $\pi_t$ against $\pi'$
       \EndFor
       \For{$\pi \in \Pi$}
         \State Update $U^\Pi_{RO}$ with expected utilities 
for games ($\pi_t$, $\pi$) 
       \EndFor   
       \If {$RO \in \{\text{win rate, normalized performance gap}\}$}
         \State $\sigma = \text{softmax}(U^\Pi_{RO})(\Pi)$ \Comment{the MSS-RO is based on utilities}
       \Else
         \State $\sigma=\texttt{Uniform}(\Pi)$ 
       \EndIf
        \If {self-play} 
          \State $\Pi = \Pi \cup {\pi_t} $
       \EndIf

     \EndFor
     \State Output current policy $\pi_t$
    \end{algorithmic}
\end{algorithm}

The opponent pool is initialized with the heuristic players, resulting in a diverse set of policies $\Pi$. At the start of the game, the expected utility matrix is empty (there are no model checkpoints for the defender yet). Thus, in the first iteration, $t_0$, we use the uniform distribution as the meta-strategy solver, MSS. During each episode of iteration $t$, an opponent strategy $\pi'$ is selected from the pool, and the defender's PPO policy $\pi_t$ learns to improve its response against it.

After each training iteration, the newly trained policy $\pi_t$ is evaluated against each opponent in the pool, to update the expected utility matrix $U^\Pi_{RO}$, where the response objective $RO$ dictates which utility function (e.g., reward, win rate by ownership or normalized performance gap) is used to compute the best response. 

For the win rate and the normalized performance gap, we define new meta-strategy solvers $\sigma$ that consider expected utilities and use softmax to transform raw scores into probabilities. To prioritize training against the most difficult opponents, we define $\sigma$ using softmax:
\begin{equation}
   \sigma = \text{softmax}(U^\Pi_{RO})(\Pi) 
\end{equation}
In the experimental section, we explore how learning improves with Flip-PSRO in two cases: (1) the opponent pool is fixed, consisting of parameterized heuristic attackers, and (2) the opponent pool is extended with PPO strategies learned at previous iterations.

\section{Experimental Results} 
\label{section:evaluation}
\noindent\textbf{\ourflip{} settings:} Throughout the experiments we consider a \ourflip{} game with $100$ time steps, $1$ resource, and $2$ players (Defender and Attacker). Each agent can choose between 3 actions: Sleep, Check and Flip. These actions have a cost of $0.0, 1.0, \text{and } 2.0$, respectively, where the high cost of Flip ensures that the agent does not attempt to claim the resource at every turn. The players gain $+1$ for each time step of resource ownership. Although the experiments in this paper focus on the single-resource setting, our \ourflip{} implementation is versatile to both single-resource and the multi-resource scenarios. We leave the study of multi-resource settings for future work.

\vspace{2pt}
\noindent\textbf{Default heuristic parameters:} The heuristic agents use the following parameters, unless otherwise specified: phase 4, random delay, burst length 3 (and phase 8 for Burst only), awakening rate $0.05$. An ablation study on other parameters is also discussed, as well as model transfer to new variants.

\vspace{2pt}
\noindent\textbf{Hyper-parameters in PPO:} The PPO algorithm uses Adam Optimizer with a learning rate of $0.001$, a discount factor of $\gamma=0.99$, clipping parameter of $\epsilon=0.2$, four epochs per agent updates, with updates every $10$ episodes.
For training we used the mean squared error and an entropy regularization coefficient of $0.01$.

\vspace{2pt}
\noindent \textbf{Flip-PSRO settings:} 
We train a PPO agent against a fixed pool consisting of heuristic agents, namely Periodic, Burst, Awakening, PC, and PAC with the default parameters. 
The learning process is carried out for 200 epochs (iterations), where each epoch lasts 100 episodes, with 10 episodes stored in memory between updates. The evaluation results for learned policies are averaged over 100 episodes.

We explore several research questions.

\begin{figure}[t]
    \centering
    \includegraphics[width=0.7\linewidth]{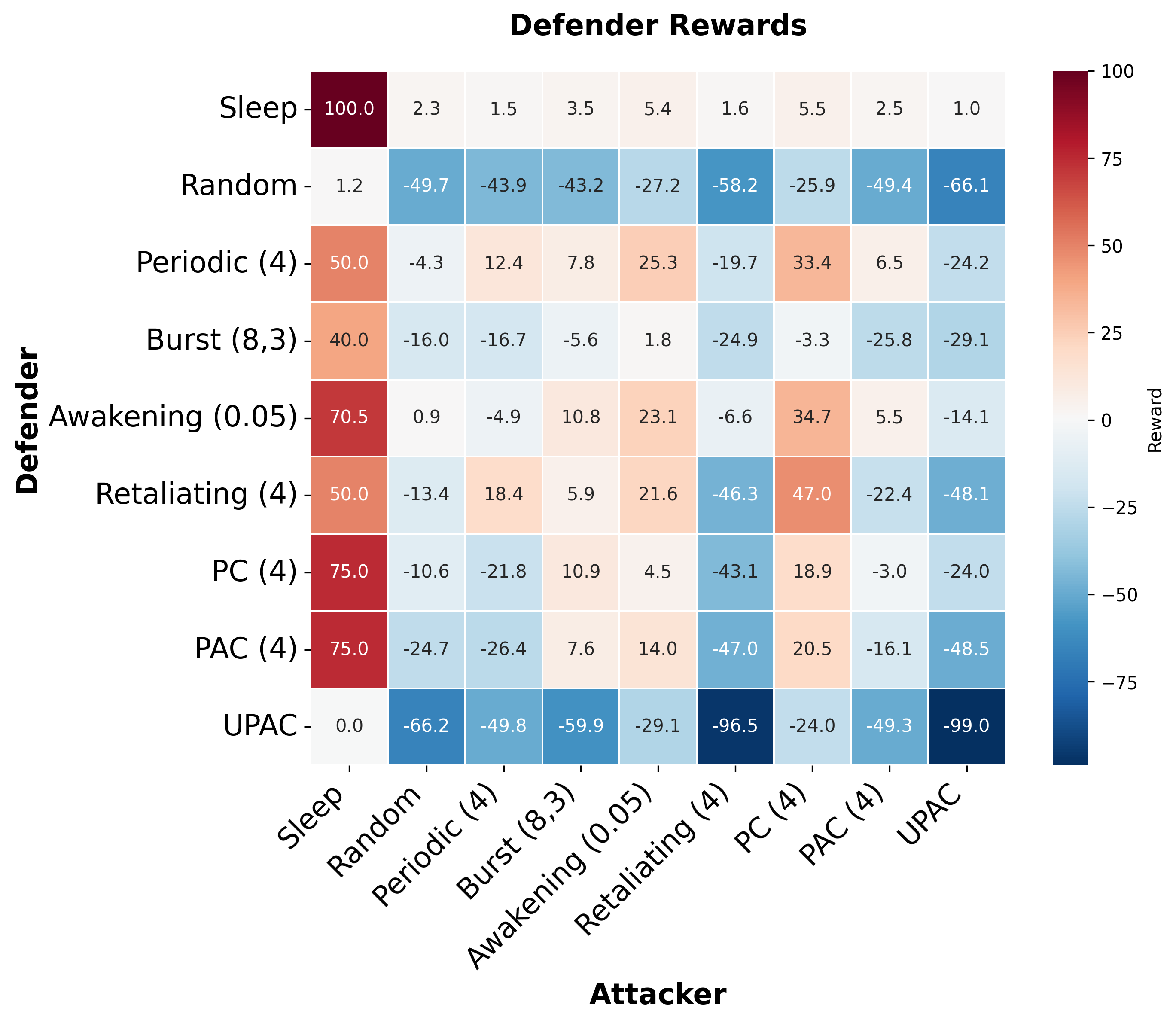}
    \caption{Defender Reward (averaged over 100 episodes), where a heuristic defender is playing against a heuristic attacker.}
    \vspace{-15pt}
    \label{fig:rew-heur}
\end{figure}

\vspace{5pt}
\noindent \textbf{(Q1) How do the different heuristic agents compare?} 
We start our evaluation with the rule-based agents playing against each other with the aim of understanding the characteristics of a well-performing heuristic. 
Figure~\ref{fig:rew-heur} presents the reward accumulated by the defender at the end of the game. Note that the game is not strictly symmetric due to the initial ownership of the resources, as the defender always starts with the resource, as expected in a real setting.
The Sleep Agent and the Random Agent (probability of flipping $33\%$) serve as baselines.  

We observe that Awakening, Retaliating and Periodic are performing generally well in the setting, in which Check has a $2\times$ lower cost compared to Flip. The strategies that use the Check action, such as Periodic Check and PAC, incur an additional cost, which decreases the reward. We also experimented with other costs for the checking action to understand how this cost affects the strategy. Interestingly, at a $20\times$ lower cost of the Check action, the best strategy becomes PAC(1), namely the periodic aggressive check strategy with a phase of 1, where the agent verifies who has control of the resource at every step. Thus, when Check is cheap, the best defense strategy is frequent investigation followed by prompt recovery. These experiments illustrate that the heuristic agents are usually not versatile because their performance is highly dependent on the testing settings. In the next experiments, we explore trained agents and their ability to generalize better than the heuristic players.

In Table~\ref{tab:ablation} we present an \textbf{ablation study} for some of the heuristic agents that perform better: Awakening, Retaliating, and PAC. We notice that the Awakening defender becomes weaker as the exponential coefficient increases. For PAC, a decrease in phase also generally reduces performance, possibly due to the more frequent costly checks.

\begin{table}[t!]
\scriptsize
\centering
\caption{Heuristic vs Heuristic with parameter sweep}
\label{tab:ablation}
\begin{tabular}{lcccccccc}
\toprule
D \textbackslash A & Awake(0.05) & Awake(0.5) & Reta(2) & Reta(4) & Reta(8) & PAC(2) & PAC(4) & PAC(8) \\
\midrule
Awake(0.05) & $23.1_{\pm 8.9}$ & $-10.2_{\pm 5.3}$ & $-21.8_{\pm 3.2}$ & $-6.6_{\pm 4.4}$ & $10.0_{\pm 6.4}$ & $-6.8_{\pm 2.1}$ & $5.5_{\pm 4.1}$ & $23.8_{\pm 5.8}$ \\
Awake(0.5) & $15.1_{\pm 4.6}$ & $-18.2_{\pm 7.7}$ & $-53.4_{\pm 4.7}$ & $-29.1_{\pm 5.2}$ & $5.9_{\pm 5.3}$ & $-21.2_{\pm 2.6}$ & $-7.5_{\pm 3.9}$ & $10.2_{\pm 3.9}$ \\
Reta(2) & $-14.5_{\pm 3.9}$ & $-29.1_{\pm 5.3}$ & $-100.0_{\pm 0.3}$ & $-3.0_{\pm 1.3}$ & $-5.1_{\pm 1.3}$ & $-65.8_{\pm 1.0}$ & $-49.0_{\pm 1.0}$ & $-24.4_{\pm 0.8}$ \\
Reta(4) & $21.6_{\pm 6.2}$ & $-14.1_{\pm 8.3}$ & $-95.4_{\pm 1.2}$ & $-46.3_{\pm 1.1}$ & $23.1_{\pm 24.8}$ & $-46.5_{\pm 0.5}$ & $-22.4_{\pm 2.0}$ & $16.6_{\pm 0.9}$ \\
Reta(8) & $27.8_{\pm 7.2}$ & $-18.0_{\pm 5.7}$ & $-43.2_{\pm 0.9}$ & $-23.5_{\pm 17.4}$ & $1.0_{\pm 10.0}$ & $-9.9_{\pm 9.2}$ & $7.3_{\pm 14.3}$ & $21.2_{\pm 4.1}$ \\
PAC(2) & $-0.3_{\pm 5.9}$ & $-60.6_{\pm 5.6}$ & $-100.1_{\pm 0.9}$ & $-94.8_{\pm 2.0}$ & $-27.2_{\pm 9.0}$ & $-69.5_{\pm 24.0}$ & $-37.4_{\pm 12.2}$ & $7.3_{\pm 6.2}$ \\
PAC(4) & $14.0_{\pm 7.1}$ & $-42.5_{\pm 5.3}$ & $-73.6_{\pm 0.9}$ & $-47.0_{\pm 1.5}$ & $-20.5_{\pm 11.9}$ & $-35.4_{\pm 12.0}$ & $-16.1_{\pm 19.8}$ & $22.3_{\pm 13.3}$ \\
PAC(8) & $13.4_{\pm 7.8}$ & $-22.6_{\pm 3.9}$ & $-36.8_{\pm 1.4}$ & $-14.2_{\pm 1.2}$ & $1.1_{\pm 3.1}$ & $-16.8_{\pm 6.1}$ & $-2.8_{\pm 12.8}$ & $18.6_{\pm 19.9}$ \\
\bottomrule
\vspace{-20pt}
\end{tabular}
\end{table}

\vspace{5pt}
\noindent \textbf{(Q2) Does a Flip-PSRO defense generalize over a population of opponents?} 
In this section, we explore the effectiveness of Flip-PSRO in learning a single policy that performs well across a population of known opponent strategies. 
We also analyze how Flip-PSRO compares with other learning paradigms described next. 

\begin{figure}[t!]
\centering
\begin{subfigure}{.45\textwidth}
  \centering
  \includegraphics[width=\linewidth]{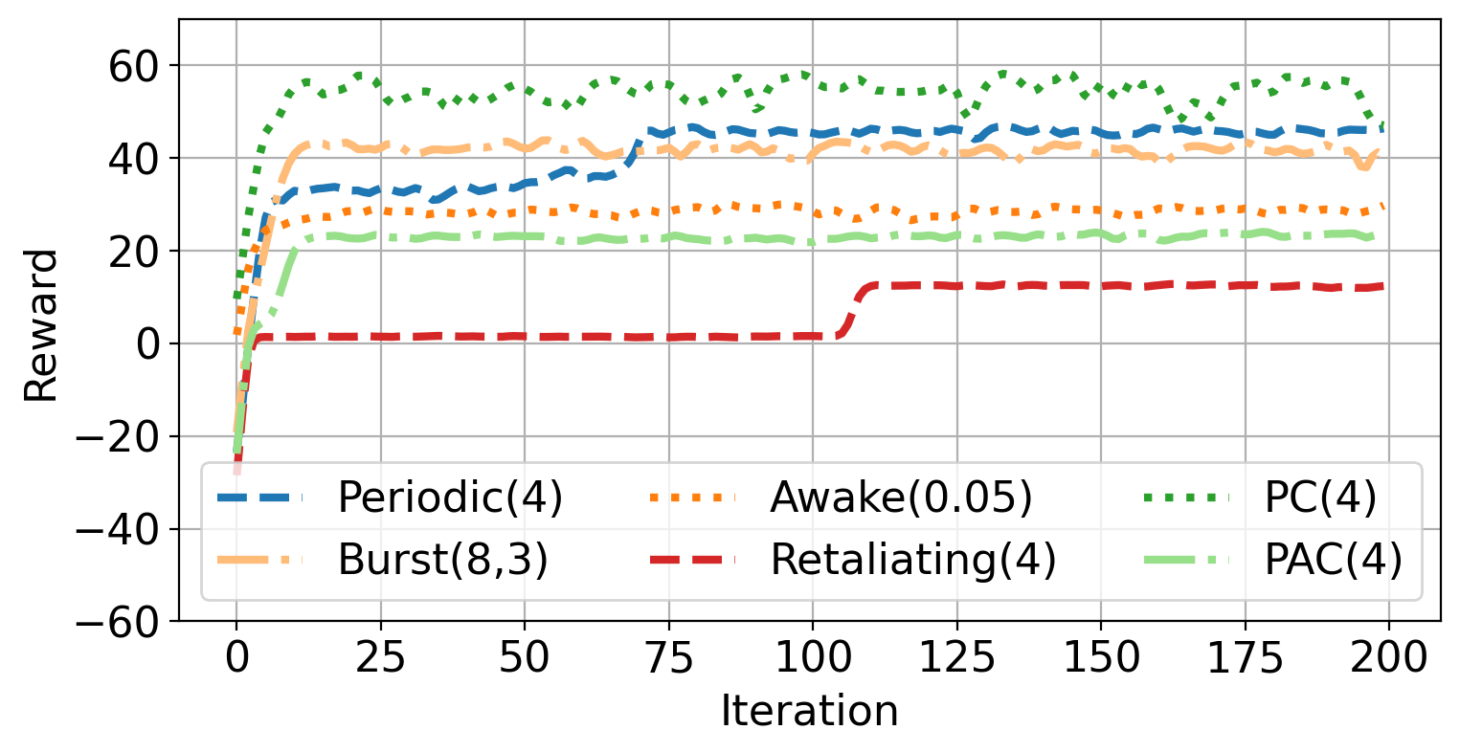}
  \caption{Specialists}
  \label{fig:spec}
\end{subfigure}
\begin{subfigure}{.45\textwidth}
  \centering
  \includegraphics[width=\linewidth]{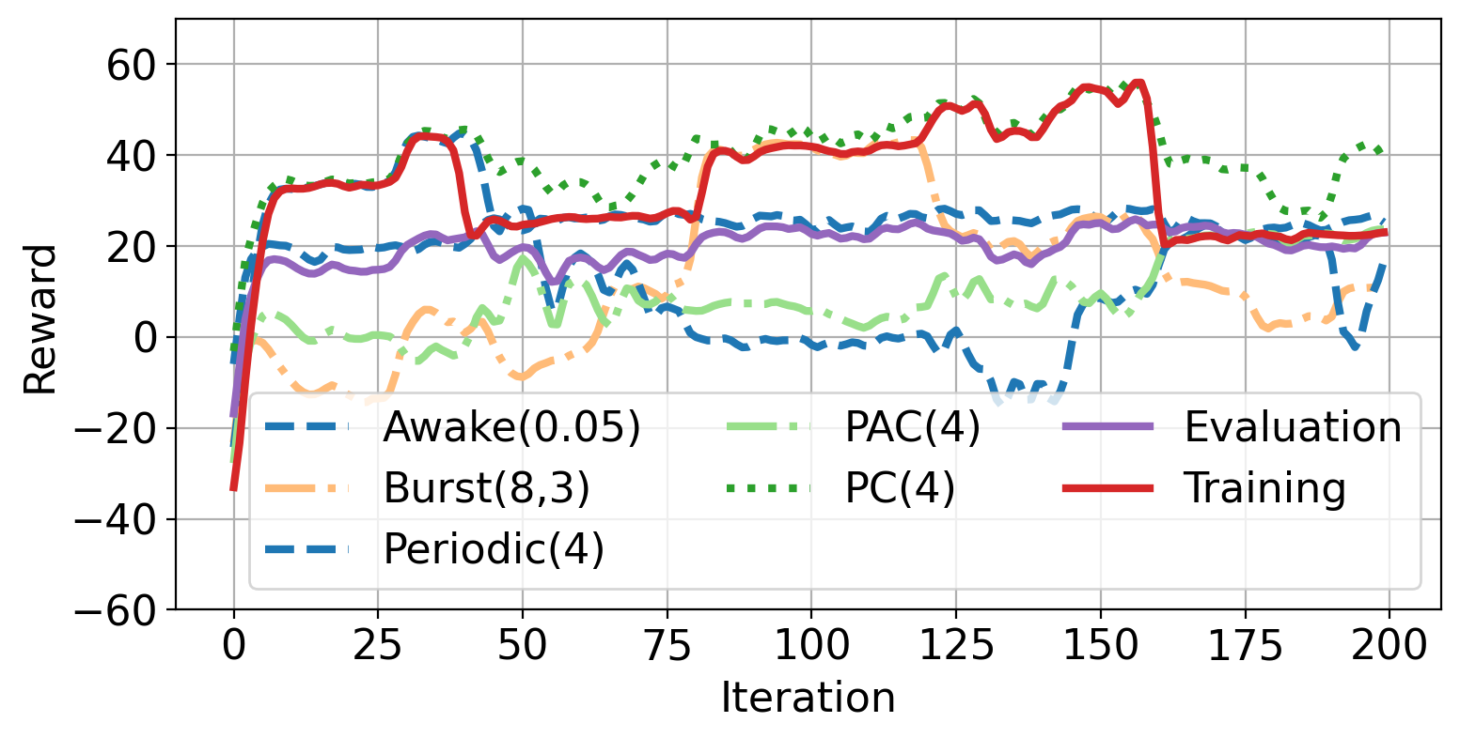}
  \caption{IBR}
  \label{fig:ibr}
\end{subfigure}

\begin{subfigure}{.45\textwidth}
  \centering
  \includegraphics[width=\linewidth]{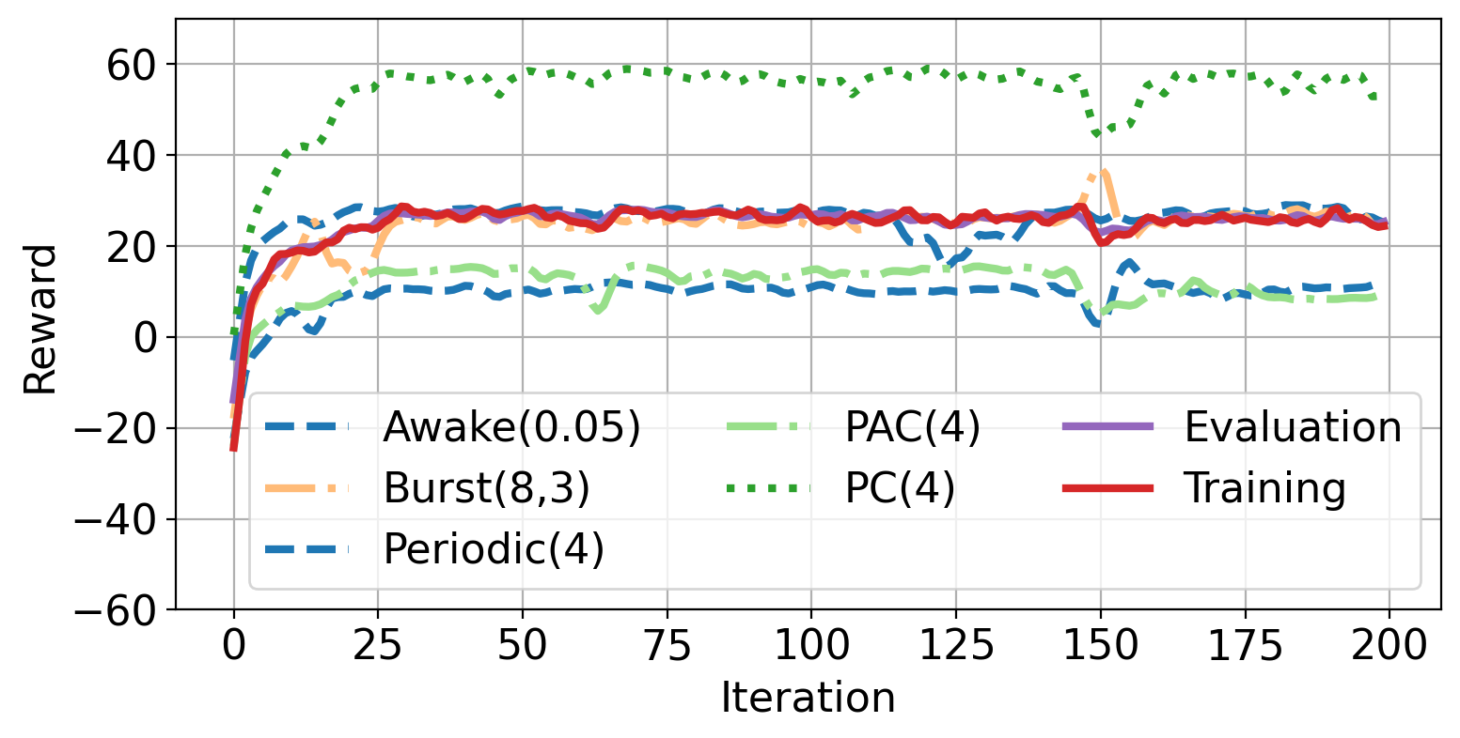}
  \caption{MSS-Uniform}
  \label{fig:mss-unif}
\end{subfigure}%
\begin{subfigure}{.45\textwidth}
  \centering
  \includegraphics[width=\linewidth]{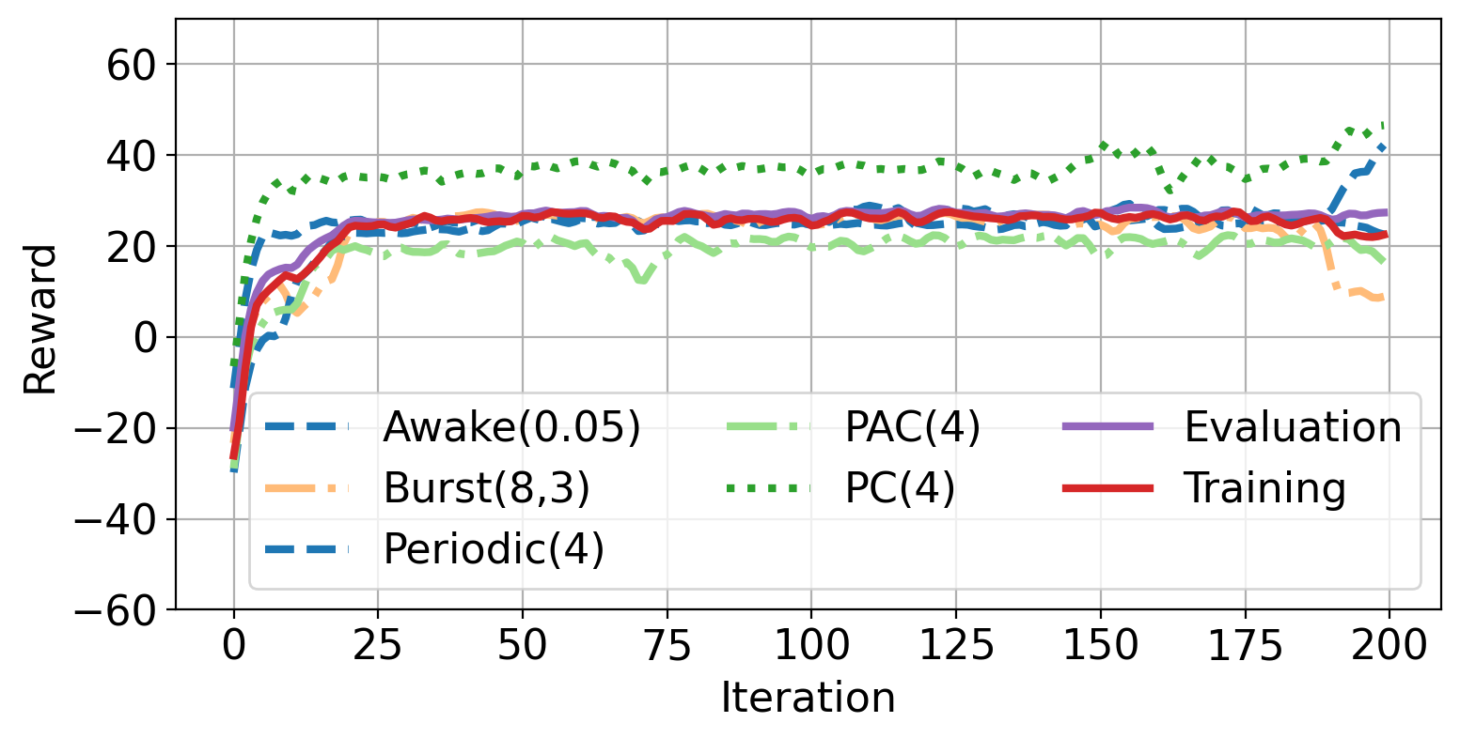}
  \caption{MSS-Ownership 70\%}
  \label{fig:mss-own70}
\end{subfigure}
\caption{Episode rewards for the trained agent (defender) against various strategies from the pool, as well as the average reward over all the strategies.
}
\vspace{-15pt}
\label{fig:tpsro}
\end{figure}

\vspace{3pt}
\noindent \textbf{Baselines:}
The first baseline is represented by Specialists, which are an approximate best-response policy learned against each heuristic individual, hence an upper bound. However, training individual specialized agents becomes infeasible as the number of opponents increases or when identifying the behavior pattern during deployments is challenging.

The second baseline policy uses the Iterated Best Response (IBR) algorithm~\cite{vinyalsStarCraft2019}, 
a sequential optimization approach in which the defender computes a best response against each opponent individually, instead of distributed training over population members (as in PSRO). Therefore, IBR is an instance of PSRO, with a probability vector over the set of strategies of $[1.0, 0.0, \cdots , 0.0]$, then $[0.0, 1.0, \cdots , 0.0]$, $\cdots$, iteratively. Unlike PSRO, which performs gradual smaller policy updates, IBR typically makes large policy changes (jumping to the new best response) that tend to overfit to the policies of other players. We implement IBR using PPO, with the same hyper-parameters specified in the beginning of the experimental section, training against the pool heuristics in this order: Awakening, Burst, Periodic, PC, PAC.
The third baseline is a rule-based strategy, namely the best performing heuristic agent from Figure~\ref{fig:rew-heur}, Awakening(0.05). 

\vspace{3pt}
\noindent \textbf{Results:} 
We study three MSS functions used to guide the Flip-PSRO algorithm: (1) MSS-Unif, the uniform distribution, (2) MSS-Gap, an MSS whose response target is the normalized performance gap (Section~\ref{section:response_objectives}), and (3) MSS-Ot\%, an MSS where ownership of more than $t\%$ of the time steps is considered a win. 
Figure~\ref{fig:tpsro} shows the episodic rewards during the Specialist training, the IBR training and the Flip-PSRO training. Both MSS-Gap and MSS-O70\% prioritize training against more challenging opponents (such as Periodic(4) and PAC), resulting in a better overall defense compared to MSS-Unif. We also note the less stable training behavior of IBR, due to the large policy updates, in contrast to the continuous improvement of the Flip-PSRO learning.

Table~\ref{tab:psro-reward} presents the accumulated rewards of the defender when playing against each of the adversaries in the pool, as well as the mean reward over all of them. The Specialist serves as an upper-limit baseline, with an average reward of 40. We observe that MSS variants reach an average reward of $\approx 27-31$, higher than the best performing heuristic, i.e., Awakening(0.05), or the Iterated Best Response, which score 13.8 and 23.2 on average, respectively. MSS-Gap prioritizes the sampling of opponents that are further from the corresponding Specialist, leading to a performance increase over the uniform pool sampling.  In addition, we illustrate two ownership targets, 50\% and 70\%, and observe that prioritizing by ownership is implicitly beneficial for the reward, as controlling the resource increases the gain.
\begin{table}[t!]
\centering
\caption{The reward of the defender trained with Flip-PSRO, compared to the baselines.}
\label{tab:psro-reward}
\begin{tabularx}{\textwidth}{L|LLLLL|L}
\hline
     & Periodic(4) & Burst(8,3) & Awake(0.05) & PC(4) & PAC(4) & Avg\\
\hline
Specialist & $46.5_{\pm 2.3}$ & $42.0_{\pm 4.8}$ &$30.1_{\pm 7.5}$ & $57.8_{\pm 3.8}$ & $23.7_{\pm 2.2}$ & $40.0$ \\
\hline
Awake(0.05) & $-4.9_{\pm 5.4}$ & $10.8_{\pm 7.8}$ & $23.1_{\pm 8.9}$ & $34.7_{\pm 5.5}$ & $5.5_{\pm 4.1}$ & $13.8$ \\
IBR & $21.9_{\pm 3.5}$ & $9.2_{\pm 2.7}$ & $25.3_{\pm 6.3}$ & $37.1_{\pm 3.6}$ & $23.5_{\pm 1.9}$ & $23.2$ \\ 
\hline
MSS-Unif & $11.1_{\pm 2.8}$ & $26.7_{\pm 2.9}$ & $25.0_{\pm 7.1}$ & $56.9_{\pm 3.7}$ & $9.3_{\pm 2.4}$ & $25.8$ \\
MSS-Gap & $24.9_{\pm 8.2}$ & $26.7_{\pm 1.6}$ & $24.8_{\pm 7.3}$ & $35.7_{\pm 4.8}$ & $22.2_{\pm 2.8}$ & $28.9$ \\
MSS-O50\% & $36.8_{\pm 10.5}$ & $27.1_{\pm 3.3}$ & $27.6_{\pm 7.1}$ & $47.4_{\pm 3.4}$ & $16.4_{\pm 2.7}$ & $31.1$ \\
MSS-O70\% & $43.5_{\pm 5.0}$ & $9.3_{\pm 2.1}$ & $22.3_{\pm 6.2}$ & $46.6_{\pm 1.9}$ & $15.4_{\pm 2.2}$ & $27.4$ \\
\hline
\end{tabularx}
\vspace{-15pt}
\end{table}

\vspace{5pt}
\noindent \textbf{(Q3) Can the Defender maintain a high ownership while optimizing performance?} 
A reliable system requires high availability of its resources to users. Controlling the system for only half the time may not represent a ``win'' for the defender in realistic settings. 
We further explore whether the MSS-Ownership function is able to steer the defense strategy towards a good trade-off between performance (reward) and control.
\begin{table}[t!]
\centering
\caption{The ownership time percentage of the defender trained with Flip-PSRO.}
\label{tab:psro-ownership}
\begin{tabularx}{\textwidth}{L|LLLLL|L}
\hline
     & Periodic(4) &  Burst(8,3) & Awake(0.05) & PC(4) & PAC(4) & Avg \\
\hline
MSS-Unif & $39.0_{\pm 3.1}$ & $67.7_{\pm 2.0}$ & $66.6_{\pm 8.2}$ & $85.4_{\pm 1.7}$ & $43.8_{\pm 2.2}$ & $60.5$ \\
MSS-Gap & $76.1_{\pm 7.5}$ & $69.5_{\pm 1.3}$ & $66.8_{\pm 6.7}$ & $78.3_{\pm 7.0}$ & $72.7_{\pm 3.0}$ & $72.28$ \\
MSS-O50\% & $86.6_{\pm 10.9}$ & $67.0_{\pm 4.6}$ & $71.7_{\pm 7.7}$ & $94.4_{\pm 5.2}$ & $61.8_{\pm 4.6}$ & $76.3$ \\
MSS-O70\% & $95.2_{\pm 4.2}$ & $69.9_{\pm 0.8}$ & $76.7_{\pm 5.7}$ & $97.6_{\pm 1.7}$ & $66.2_{\pm 2.0}$ & $81.12$ \\

\hline
\end{tabularx}
\vspace{-15pt}
\end{table}

Table~\ref{tab:psro-ownership} presents the defender's ownership, calculated as the percentage of time steps when the defender controls the resource. These results show that the proposed MSS is successful in acquiring a high level of control, while also achieving a high reward over the mixture of strategies (see reward from Table~\ref{tab:psro-reward}). 
As the ownership target (i.e., minimum level of control for the defender to be declared winner) increases from 50\% to 70\%, the actual average ownership across the pool members also increases from 76 to 81 (Table~\ref{tab:psro-ownership}). 

\vspace{5pt}
\noindent \textbf{(Q4) Do policies trained with Flip-PSRO transfer to unseen opponents?} 
In this section, we explore the ability of Flip-PSRO to adapt to variants of attack strategies that were not present in the training pool. 
Table~\ref{tab:psro-transfer} shows the test performance of Flip-PSRO under three meta solvers, Uniform, Ownership 50\% and Ownership 70\%, and the IBR baseline. We construct several new attack variants that were not present during training, by changing the phase and burst length. Previous work points out the importance of population diversity and opponent similarity for transfer quality; thus, transfer works best when unseen opponents fall within the convex hull of training strategies~\cite{lian2025fusionpsronashpolicyfusion,NEURIPS2023_d61819e9}
\begin{table}[t!]
\centering
\caption{Transferability to unseen opponents. The models trained on the strategies specified earlier (see Figure~\ref{fig:tpsro}) are evaluated on new attack variants.
}
\label{tab:psro-transfer}
\begin{tabular}{c|lllllll|l}
\hline
     & P(6) & P(8) & B(8, 6) & B(16,3) & PC(8) & PAC(6) & PAC(8) & Avg\\
\hline
IBR & -10.4 & 36.8 & -19.0 & 27.9 & 43.0 & -10.9 & 32.1 & 14.2 \\
\hline
MSS-Unif & 14.8 & 55.1 & -11.0 & 24.0 & 37.4 & 14.6 & 55.5 & 27.2 \\
MSS-O50\% & 29.6 & 45.9 & 4.2 & 37.2 & 49.4 & 29.9 & 30.1 & 32.3 \\
MSS-O70\% & 4.2 & 47.2 & -12.4 & 26.7 & 46.6 & 4.2 & 24.3 & 20.1 \\
\hline
\end{tabular}
\vspace{-10pt}
\end{table}

To address this caveat, we train Flip-PSRO to cover a diverse population of policies, which creates robustness against various playing styles. From Table~\ref{tab:psro-transfer}, we see that Flip-PSRO transfers best against new agents whose phase is a multiple of the training phase (of 4). Thus, the Flip-PSRO defender is successful against P(8), B(16, 3), PC(8), PAC(8), but less successful against P(6) and PAC(6). Our intuition is that multiples of phase are similar to the agent waiting for the next well matched opportunity to act. Furthermore, we observe that IBR has weaker transfer properties to unseen opponents, since its sequential best-response training makes it more susceptible to overfitting to particular opponent characteristics~\cite{vinyalsStarCraft2019}. We also observed that MSS-Gap, similar to IBR, is challenged by unseen variants. We believe that the utility function based on the training performance gap makes the policy less generalizable to new opponents, and leave detailed exploration of MSS-Gap for future work.

\begin{figure}[t]
\centering
\begin{subfigure}{1\textwidth}
  \centering
  \includegraphics[width=\linewidth]{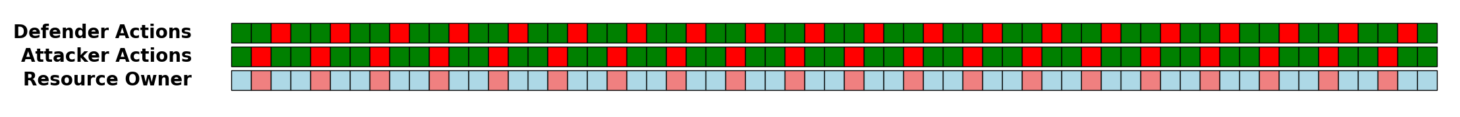}
  \caption{Variant from epoch 97 playing against variant from epoch 6}
  \label{fig:ppo-97-vs-ppo-6}
\end{subfigure}%

\begin{subfigure}{1\textwidth}
  \centering
  \includegraphics[width=\linewidth]{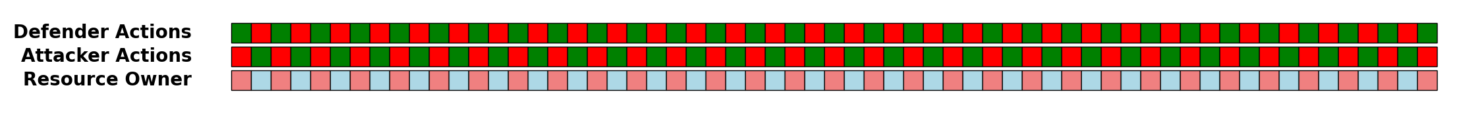}
  \caption{Variant from epoch 6 playing against variant from epoch 97}
  \label{fig:ppo-6-vs-ppo-97}
\end{subfigure}
\caption{Trained opponents playing against each other. We select an early variant produced before learning against the pool has converged (epoch 6), and one of the later variants, after convergence (epoch 97).}
\vspace{-15pt}
\label{fig:self-play-actions}
\end{figure}
\vspace{5pt}
\noindent \textbf{(Q5) How does self-play impact learning in Flip-PSRO?}
In \ourflip{}, there is a close symmetry between the Defender and the Attacker, therefore the two players may use exact same policies. In our experiments, we also explored learning through self-play, by extending the pool with earlier versions of the trained agent.
The motivation is to maintain a population of past versions to avoid forgetting good strategies~\cite{ICLR23-Cui,vinyalsStarCraft2019}.
We trained the Flip-PSRO agent against a uniform mixture of opponents from the extended pool comprised of the heuristic opponents and versions of itself (using a flip cost of 2 and a checking cost of 0.1). We observed that performance against the heuristic opponents in the pool decreases as the resulting strategy attempts to improve on previous versions of itself. The reason is that the PPO agent learns early a good (obvious) strategy that relies on checking for possession frequently and flipping once control has been lost. 
Figure~\ref{fig:self-play-actions} shows the Flip-PSRO agent playing against another version of itself. The two opponents effectively counteracted each other, resulting in a stalemate situation, where both alternate between checking and flipping once they lost control.
We believe that self-play is beneficial in more complex settings with extended action spaces, whose study we leave for future work.

\section{Conclusion} 
\label{section:conclusion}
In this work, we introduced \ourflip{}, a multi-agent reinforcement learning environment designed to extend the classic FlipIt cybersecurity game. 
We proposed Flip-PSRO, a policy-space response oracle framework tailored to train robust defenders through iterative learning against a diverse pool of attackers.
By incorporating response objectives such as win rate by ownership and normalized performance gap, Flip-PSRO allows agents to focus their training on harder-to-beat opponents.
Our comprehensive evaluation showed that RL agents can successfully learn specialized strategies that outperform heuristic adversaries and, furthermore, generalist agents trained via Flip-PSRO can successfully generalize across diverse attack patterns.
As future work, we intend to study the 2-player learning setting in Flip-PSRO in more detail, and explore convergence guarantees to an equilibrium in symmetrical and asymmetrical settings.

\subsubsection*{Acknowledgments} 
This research was funded by the Defense Advanced Research Projects Agency (DARPA), under contract W912CG23C0031.

\bibliography{references}
\bibliographystyle{unsrt}

\end{document}